\documentclass[twocolumn,letter,10pt]{IEEEtran} %!PN
\def \enTitle{\vspace{-3mm}Neural Network Ensembles to Real-time Identification of Plug-level Appliance Measurements}
\def \student{Karim Said Barsim}   % FIXME
\def \worksubject{NILM}

\def \doclang{english}
\def \colortype{boxed} 	% color, bw, boxed
\def \keywords
{
	Plug-level monitoring,
	Distributed sensing,
	Neural network ensembles,
	PLAID dataset
}
\def \paperAbstract
{
    The problem of identifying end-use electrical appliances from their individual consumption profiles, known as the appliance identification problem, is a primary stage in both Non-Intrusive Load Monitoring (NILM) and automated plug-wise metering.
    Therefore, appliance identification has received dedicated studies with various electric appliance signatures, classification models, and evaluation datasets.
    In this paper, we propose a neural network ensembles approach to address this problem using high resolution measurements.
    The models are trained on the raw current and voltage waveforms, and thus, eliminating the need for well engineered appliance signatures.
    We evaluate the proposed model on a publicly available appliance dataset from 55 residential buildings, 11 appliance categories, and over 1000 measurements.
    We further study the stability of the trained models with respect to training dataset, sampling frequency, and variations in the steady-state operation of appliances.
}

\usepackage[ruled,vlined]{algorithm2e}

\usepackage[cmex10]{amsmath}	%\usepackage{amsmath}
\usepackage{amssymb}
\usepackage{array}

\usepackage{bm}
\usepackage{caption}	%\usepackage[format=hang]{caption} 	%\usepackage[caption=false]{caption}
\usepackage[final]{changes}

\usepackage{dsfont}
\usepackage{enumitem}
\usepackage{etoolbox}
\usepackage{eucal}

\usepackage{fullpage}

\usepackage[right = 15.875mm, left = 15.875mm, top = 19mm, bottom = 25.4mm]{geometry}
\usepackage{glossaries}

\usepackage{hyperref}
\usepackage{ifthen}
\usepackage{indentfirst}

\usepackage{moresize}

\usepackage[square,numbers,sort&compress]{natbib}
\usepackage{pdfpages}
\usepackage{perpage}
\usepackage{pgfplots}

\usepackage[nodisplayskipstretch]{setspace}
\usepackage{stfloats}
\usepackage{subcaption}

\usepackage{textcomp}
\usepackage{tikz}
    \makeatletter\newcommand\subparagraph{\@startsection{subparagraph}{5}{\parindent}{3.25ex \@plus 1ex \@minus .2ex}{-1em}{\normalfont\normalsize\bfseries}}\makeatother
\usepackage[compact]{titlesec}
    \let\subparagraph\relax
\usepackage[color = yellow]{todonotes}
\usepackage{url}
\usepackage{epstopdf}
\ifthenelse{\equal{\doclang}{german}}
{
	\def \langtitle{\deTitle}
		
}
{
	\def \langtitle{\enTitle}
	
}

\pdfstringdef \studentPDF {\student} 
\pdfstringdef \worktitlePDF {\langtitle}
\pdfstringdef \worksubjectPDF {\worksubject}
\pdfstringdef \keywordsPDF {\keywords}
\hypersetup{pdfauthor=\studentPDF, 
            pdftitle=\worktitlePDF,
            pdfsubject=\worksubjectPDF,
            pdfkeywords=\keywordsPDF}
            
\usetikzlibrary{arrows.meta,
				automata,
				backgrounds,
                calc,
				decorations,
				decorations.pathmorphing,
				decorations.pathreplacing,
                intersections,
                ocgx,
				patterns,
				petri,
				positioning,
				shapes,
				spy,
				topaths}
\pgfplotsset{compat=1.13}
\titlespacing{\section}{0pt}{5pt}{2pt}
\newcommand{\argmax}{\operatornamewithlimits{arg\ max}}
\MakePerPage{footnote}

\SetAlgorithmName{Algorithm}{Algorithm}{List of algorithms}

\renewcommand{\baselinestretch}{0.97531}

\def\BibTeX{{\rm B\kern-.05em{\sc i\kern-.025em b}\kern-.08em
		T\kern-.1667em\lower.7ex\hbox{E}\kern-.125emX}}

\hypersetup{pdfstartview=FitH, bookmarks=true, bookmarksnumbered=true, bookmarksopen=false, breaklinks=false,
		    citecolor = blue}
\ifthenelse{\equal{\colortype}{color}}
	{
		\hypersetup{colorlinks = true, linkcolor = blue}
	}
	{
		\ifthenelse{\equal{\colortype}{boxed}}
		{}
		{
			\hypersetup{hidelinks=true}
		}
	}

\newcommand{\plus}{\texttt{+}}
\newcommand{\minus}{\texttt{-}}

\presetkeys{todonotes}{fancyline, linecolor = green!30, backgroundcolor=yellow!70, bordercolor = black}{}

\makeglossaries

\newglossaryentry{Classifier}{
    name = Classifier (prediction function) $\hat{f}$ :,
    description={an estimation of nominal-valued function based on external information (labeling) that maps a set of observations to a label space $f:\mathcal{X}\longmapsto \mathcal{Y}$}}

\newglossaryentry{Observation}{
    name = {Observation, sample, instance, or event $\bm{x}$ :},
    description = {the feature vector (attribute-value pairs) of a detected event (in the event based NILM system)}}

\newglossaryentry{Training data}{
    name = training data,
    description={a set of pairs each containing a feature vector of an observation and its confirmed class label}}

\newglossaryentry{Label space}{
    name = Label space $\mathcal{Y}$ :,
    description = {the set of all possible class labels (e.g. names of all underlying (known) loads).}}

\newglossaryentry{Learner}{name = Learner (learning algorithm) $h$ :,
    description={an estimator of a prediction function (classifier). A semi-supervised learner utilizes both labeled and unlabeled observation $\hat{f} = h(\mathcal{L},\mathcal{U})$ while  supervised learner estimates the prediction function merely from the labeled samples $\hat{f} = h(\mathcal{L})$}}

\newglossaryentry{Sample space}{
    name = Sample space $\mathcal{X}$:,
    description={the set of all (labeled and unlabeled) observations available to the learner}}

\newglossaryentry{Confidence-rate classification}{
    name = Confidence-rate classifier :,
    description={is a prediction function whose output is a pair of a nominal-valued label and real-valued score that indicates the confidence of the function in its prediction.}}

\begin{document}

%===============================================================================================================
%=======================  Paper heading (title, authors, abstract, and keywords)  ==============================
\title{\enTitle}	% Paper title ... can use linebreaks \\ within to get better formatting as desired

%===============================================================================================================
%====================================== Author names and affiliations ==========================================
% Use a multiple column layout for up to three different affiliations
\author
{
 \vspace{3mm}
 \IEEEauthorblockN{Karim Said Barsim, Lukas Mauch, and Bin Yang\\}
 \IEEEauthorblockA{Institute of Signal Processing and System Theory, University of Stuttgart\\
				   \texttt{\{karim.barsim,lukas.mauch,bin.yang\}@iss.uni-stuttgart.de}}
 \vspace{-3mm}
}

%===============================================================================================================
%============================================ Make the title area ==============================================
\maketitle

%===============================================================================================================
%============================================= Make the abstract ===============================================
\begin{abstract}\paperAbstract\end{abstract}

%===============================================================================================================
%============================================= Make the keywords ===============================================

%===============================================================================================================
%========================== ****************  Main content  **************** ===================================
\section{Introduction}
\label{sec:introduction}

The energy end-use monitoring problem has received a widespread attention over the last two decades, either in a non-intrusive fashion using smart meters which aim at reducing the cost of installation and maintenance \cite{Hart_1992,Froehlich_2011} or distributed sensing and smart outlets that became available through recent technological developments \cite{Barker_2014,Gao_2015}. 
Both trends face what is referred to as the appliance identification problem. 
In non-intrusive monitoring, the disaggregation stage reconstructs end-appliance consumption profiles from the aggregate measurements in a building. 
Afterwards, the appliance identification stage assigns either a generic end-use category \cite{Parson_2014} or specific appliance instances \cite{Kolter_2012_Approximate} to each reconstructed load profile.
In plug-level metering, a smart outlet monitors the real-time consumption of its plugged-in appliance but this appliance may change over time based on user requirements.
Similarly, identifying the connected appliance from the given consumption profile without user intervention is the challenge of appliance identification.

Previous works have addressed the problem of appliance identification using various machine learning tools, electric signatures, or evaluation platforms. 
In \cite{Barker_2014} for instance, \underline{s}upport \underline{v}ector \underline{m}achines (SVM) and \underline{d}ecision \underline{t}rees (DTree) trained on three-month, second-based power profiles of over 20 appliances achieved comparable performance on future runs of the same appliances. The performance degraded notably as previously unseen appliances were added to the test.
In \cite{Reinhardt_2012_Electric}, current harmonics in addition to a few more features were extracted from high-resolution current waveforms (during transient and steady-state operations) to compare various classification algorithms.
High accuracies were reported using Bayesian networks evaluated on 16 categories of appliances and over 3000 acquired measurements.
A more recent study, and most related to this work, compared and evaluated nine classifiers (including kNNs, logistic regression, SVMs, and DTrees) on different raw and engineered features \cite{Gao_2015} from a publicly available power dataset \cite{Gao_2014}. The authors proposed a novel appliance signature based on quantization of the VI-trajectories into a binary VI-image.
The study further examined the effect of the sampling frequency on the identification problem showing that high classification rates can be obtained using random forest trees trained on the raw current and voltage measurement of at least 4kHz of sampling frequency (up to 82\% of accuracy using VI-images and 86\% using a set of combined features).

In most of these works neural networks on raw measurements have received little or no attention as a candidate model for appliance identification.
Neural networks have recently received a growing interest in various machine learning applications especially after the evaluation of a deep convolutional architecture on a visual recognition dataset known as ImageNet.
Recently, different deep learning architectures were found to achieve comparable results in the problem of energy disaggregation \cite{Kelly_2015,Mauch_2015}. 

In this paper, we first complement the work in \cite{Gao_2015} with an ensemble of neural networks in addressing the appliance identification problem from the raw, high-resolution current and voltage waveforms.
We further study the stability and robustness of the proposed model with respect to the size of the training dataset, the sampling frequency of the measurements, and signal variations during the steady-state operation of appliances.
We evaluate our models on the publicly available \underline{p}lug \underline{l}oad \underline{a}ppliance \underline{i}dentification \underline{d}ataset (PLAID) \cite{Gao_2014} which contains, at the time of this work, measurements from 11 appliance categories, 235 appliance instances distributed over 55 households, and a total of 1074 current and voltage measurements at a sampling frequency $f_s=30$kHz and a grid frequency $f_g=60$Hz.

\section{Appliances' signatures}
\label{sec:signatures}

In this work, we utilize the raw current $i(n)$ and voltage $v(n)$ waveforms during the steady-state operation as an appliance signature. 
%\autoref{fig:vitrajectories} shows various current, voltage, and VI-trajectories of different appliance categories from the PLAID dataset.
%The selected samples show possible discrimination between different categories using these signatures.
Unlike previous works on VI-trajectories that extracted shape-based features \cite{Lam_2007,Hassan_2014}, we exploit one of the advantages of neural nets by utilizing the raw signals directly.

Given current $i(n)$ and voltage $v(n)$ signals of an appliance at a sampling frequency $f_s$ and a grid frequency $f_g$, we extract one complete period of each waveform 
\vspace{-1.75mm}
\begin{equation*}
\bm{\varphi}_\tau =  \big[\varphi(\tau), \;\varphi(\tau\plus 1), \;\dots\;,\;\varphi(\tau\plus d\minus 1)  \big]^T \;\;\in \mathbb{R}^d,\;\; \varphi \in \{i, v\}
\end{equation*}
where $d = f_s/f_g$ is the number of samples per period and $\tau$ is a point in time during the steady-state operation of the appliance. In order to support category-based appliance identification, we alleviate all amplitude information through segment-based normalization where each extracted signal segment $\bm{\varphi}_\tau$ is normalized to the range [-1,1]. The result is the signal segments
\vspace{-1mm}\begin{equation}\vspace{-1mm}
\hat{\bm{\varphi}}_\tau =  \frac{2\bm{\varphi}_\tau - \big(\max \bm{\varphi}_\tau + \min \bm{\varphi}_\tau \big) \bm{J}_{d,1}}{\big(\max \bm{\varphi}_\tau - \min \bm{\varphi}_\tau \big)} \quad \in [-1,1]^{d}
\end{equation}
where $\bm{J}_{m,n}$ is an $m$-by-$n$ matrix of ones. The input vector to the neural network becomes
\vspace{-1.75mm}\begin{equation}\vspace{-1.25mm}
\bm{x} = \Big[\;\hat{\bm{i}}_\tau^T\;,\;\; \hat{\bm{v}}_\tau^T \;\Big]^T \quad \in [-1,1]^{2d}
\end{equation}
which corresponds to an input layer of size $2d$ for each neural network.
We further expand the size of the training set algorithmically through extracting several signal segments from equally spaced initial phases of each measurement
\vspace{-1.25mm}\begin{equation}\vspace{-1mm}
\mathcal{X} =  \left\{\;\bm{x}_\tau \;\Big|\; \tau = \tau_o\plus i\, \varepsilon\, ,\;0\leq i < \frac{d}{\varepsilon} \right\}
\end{equation}
where $\varepsilon$ is the sliding step of the extraction window and this is applied to each measurement. Algorithmic expansion of the training data has several advantages. It provides a larger set of training data, eliminates the need for phase alignment of the signals, and more importantly drives a fully connected neural network to become invariant to the initial phase of the extracted signals. As a result, the model becomes more robust to signal variations during steady-state operation of appliances.
\section{Prediction model}
\label{sec:prediction_model}

Given a set $\Omega = \{\omega_m\}_{m=1}^{M}$ of $M$ class labels (i.e. appliance categories), the straightforward approach for appliance classification is a multi-class neural network. However, in our initial tests we observed that an ensemble of binary networks performs consistently better than the multi-class model with a similar total size (i.e. neurons), training algorithm, regularization method, and overall training time. It is known that neural networks are one of the unstable learning algorithms with respect to training data \cite{Aggarwal_2015} (i.e. they are sensitive to changes in the training set). For unstable models, an ensemble approach known as Bootstrap aggregation (aka Bagging) \cite{Breiman_1996,Friedman_2007} is expected to provide better results than a single base model\footnote{In Bootstrap aggregation, an ensemble of weak, multi-class learners, each trained on a randomly sampled subset of the whole training dataset, is used rather than binary-classifiers generalized to a multi-classification problem.}. We, therefore, adopt the latter approach (i.e. an ensemble of binary networks) and discuss it further in the following.

For each class combination $(\omega_i,\omega_j)_{i<j}$, a binary classification neural network $\hat{\bm{\theta}}_{\omega_i,\omega_j}$ is trained on an appropriate subset of the training data
\begin{equation*}
\mathcal{D}_{\omega_i,\omega_j} = \left\{ \bm{x} \in \mathcal{X} \;\big|\;  \omega(\bm{x})  = \omega_i \;\text{or}\; \omega(\bm{x})  = \omega_j  \right\} \quad \forall(i < j)
\end{equation*}
where $\omega(\bm{x})$ is the true class of the sample $\bm{x}$. Therefore, we train a total of ${{M}\choose{2}}$ base models each can be described as
\vspace{-1mm}\begin{equation}
\hat{\bm{\theta}}_{\omega_i,\omega_j}: \mathcal{X} \longmapsto [0,1]^2 \qquad \forall(i< j)
\end{equation}
\begin{equation}\vspace{-1mm}
\hat{\bm{\theta}}_{\omega_i,\omega_j}(\bm{x}) = \big[ \hat{p}_{\omega_i,\omega_j}(\bm{x}),\; \hat{p}_{\omega_j,\omega_i}(\bm{x}) \big]
\end{equation}
where $\hat{p}_{\omega_i,\omega_j}$ is an estimated normalized score for the class $\omega_i$ over $\omega_j$ for the given sample $\bm{x}$.
Assuming a normalized output of the network
\vspace{-2.5mm}\begin{equation}\vspace{-1mm}
\hat{p}_{\omega_i,\omega_j}(\bm{x}) = 1 - \hat{p}_{\omega_j,\omega_i}(\bm{x})\; \forall\bm{x},
\end{equation}
the target vector for the training of $\hat{\bm{\theta}}_{\omega_i,\omega_j}(\bm{x})$ becomes [1,0] if $\omega(\bm{x}) = \omega_i$ or [0,1] otherwise.
The final ensemble prediction function is defined as
\begin{equation}\label{eq:votinga}
\hat{\omega}(\bm{x}) = \argmax_{\omega_i\in\,\Omega} \;\sum_{j\neq i} \hat{p}_{\omega_i,\omega_j}(\bm{x})
\end{equation}
for a confidence-weighted voting or
\begin{equation}\label{eq:votingb}
\hat{\omega}(\bm{x}) = \argmax_{\omega_i\in\,\Omega} \;\sum_{j\neq i}\;\mathds{1}\Big(\hat{p}_{\omega_i,\omega_j}(\bm{x}) > \hat{p}_{\omega_j,\omega_i}(\bm{x})\Big)
\end{equation}
for unweighted majority voting where $\mathds{1}$ is the indicator function.

%In each case, several models are trained on the same dataset and one is selected for final prediction based on some selection criteria
%\vspace{-2mm}\begin{equation}
%\hat{\bm{\theta}}_{\omega_i,\omega_j} = \text{ SELECT } \left\{ \hat{\bm{\theta}}_{\omega_i,\omega_j}^{(n)} \right\}_{n=1}^{N_n}
%\end{equation}

Finally, we utilize a validation-based early stopping as a regularization technique to avoid over fitting \cite{Prechelt_1997}. As mentioned earlier, PLAID measurements contains, for each appliance category, several appliance instances from various households. In order to improve generalization ability to new buildings, a 30\% validation set is selected in a building-based fashion. 

%In other words, if the samples are distributed over the set of houses $\mathcal{H} = \{h_k\}_{k=1}^K$, a 30\% validation set means that the training set for each model $\mathcal{D}_{\omega_i,\omega_j}$ is partitioned into two subsets  
%\begin{align}
%\mathcal{D}_{\omega_i,\omega_j}^{(r)} &= \left\{\bm{x}\in \mathcal{D}_{\omega_i,\omega_j}\;\big| \; h(\bm{x}) \in \mathcal{H}^{(r)} \right\} \\
%\mathcal{D}_{\omega_i,\omega_j}^{(v)} &= \left\{\bm{x}\in \mathcal{D}_{\omega_i,\omega_j}\;\big| \; h(\bm{x}) \in \mathcal{H}^{(v)} \right\}
%\end{align}
%for training and validation where the two sets $\mathcal{H}^{(r)}$ and $\mathcal{H}^{(v)}$ are partitioning $\mathcal{H}$ such that the size of the of the later is $0.3|\mathcal{H}|$.

\section{Evaluation}

    \setlength{\belowcaptionskip}{-7mm}
\begin{figure*}
	\vspace{-2.25mm}
    \setlength{\abovecaptionskip}{-3pt}
    \centering\input{figures/VITrajectories.tex} 
    \caption{\small{Current, voltage, and VI-trajectories of selected samples from PLAID dataset \cite{Gao_2014} are shown to the left. The appliance categories and IDs, from top to bottom, are (1) air conditioner, 1010, (2) CFL, 20, (3) fan, 766, (4) fridge, 6, (5) hairdryer, 444, (6) heater, 716, (7) bulb, 57, (8) laptop 28, (9) microwave, 10, (10) vacuum cleaner, 730, and (11) washing machine, 488. The per-category evaluation metrics are shown to the right while the per-house metrics are in the bottom. The best evaluation results are $\kappa = 0.882$ and $\alpha = 0.897$.}}
    \label{fig:vitrajectories}
\end{figure*}

We evaluate our models on the publicly available PLAID dataset \cite{Gao_2014} which contains $M=11$ appliance categories. Therefore, we have ${{M}\choose{2}}=55$ class combinations and for each we train a two-layer, fully connected, feed-forward neural network with $2d$ input neurons, 30 hidden neurons, and two output neurons. Since some measurements contain turn-on transients, which are to be avoided in this work, the training set is extracted from the last two periods of each measurement. With a sliding step of $\varepsilon = 10$ samples, the training dataset is expanded by a factor of $d/\varepsilon = 50$. The hyperbolic tangent (\emph{tanh}) activation is utilized in the hidden layer while the normalized exponential (\emph{softmax}) is selected for the output layer. We utilized \texttt{matlab}'s implementation of the conjugate gradient descent with random restarts \cite{Powell_1977} as a training function for all models.
%The total training time of all 55 models is $1\sim2$ hours on a 12GB-GPU unit.

In order to compare this approach with the previously studied models on this dataset \cite{Gao_2015}, we adopt the same \emph{leave-house-out} cross validation technique. In other words, samples from one house is saved for test while the remaining set of buildings are available for training, resulting in a total of 55 test cases.

    \setlength{\belowcaptionskip}{-7mm}
\begin{figure*}
	\vspace{-2.75mm}
    \setlength{\abovecaptionskip}{0pt}
    \centering\input{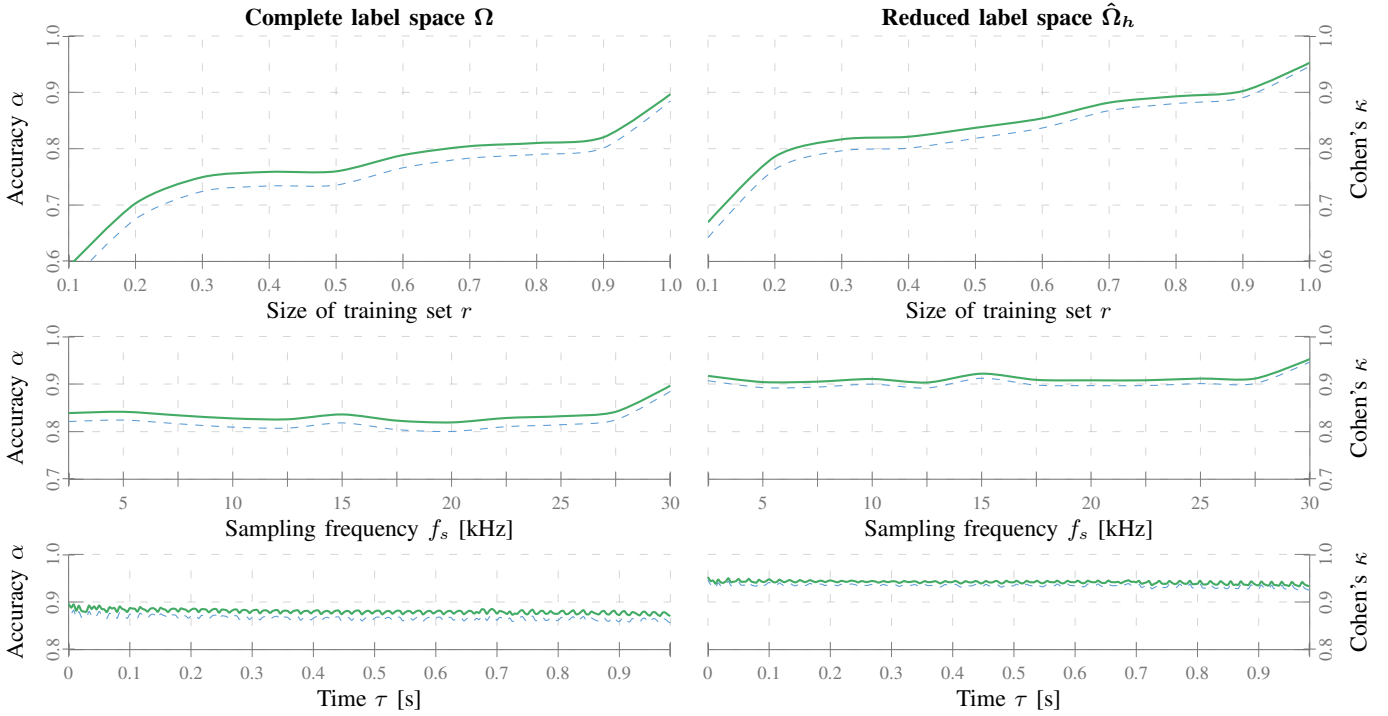} % \caption{Phase A} \label{fig:selfLearningOverTimePhaseA}
    \caption{\small{Aggregate evaluation results as a function of (a) training set size, (b) sampling frequency, and (c) time shift. To right are the same experiments with reduced label spaces based on the \emph{prior knowledge} of the label space of each target household.}}
    \label{fig:stability}
\end{figure*}

For each class $\omega_m$ and the total confusion matrix $\bm{\Lambda}=[\Lambda_{ij}]_{i,j=1}^{M}$, we introduce the following performance indicators 
{
    \small
    \begin{align}
    \text{TP}_m 			& = \left|\left\{ \bm{x} \in \tilde{\mathcal{X}} \;\big|\; \hat{\omega}(\bm{x})   =  \omega_m \text{ \& } \omega(\bm{x})   =  \omega_m  \right\}\right| \\
    \text{TN}_m 			& = \left|\left\{ \bm{x} \in \tilde{\mathcal{X}} \;\big|\; \hat{\omega}(\bm{x}) \neq \omega_m \text{ \& } \omega(\bm{x}) \neq \omega_m  \right\}\right| \\
    \text{FP}_m 			& = \left|\left\{ \bm{x} \in \tilde{\mathcal{X}} \;\big|\; \hat{\omega}(\bm{x})   =  \omega_m \text{ \& } \omega(\bm{x}) \neq \omega_m  \right\}\right| \\
    \text{FN}_m 			& = \left|\left\{ \bm{x} \in \tilde{\mathcal{X}} \;\big|\; \hat{\omega}(\bm{x}) \neq \omega_m \text{ \& } \omega(\bm{x})   =  \omega_m  \right\}\right| 
    \end{align}
    \vspace{-5mm}
    \begin{alignat}{5}
    \text{recall}_m        &= \text{TPR}_m & &= \text{TP}_m  &          &/\; ( \text{TP}_m  &        &+ \text{FN}_m  &          &)           \\
    \text{precision}_m     &= \text{PPV}_m & &= \text{TP}_m  &          &/\; ( \text{TP}_m  &        &+ \text{FP}_m  &          &)           \\
    \text{specificity}_m   &= \text{TNR}_m & &= \text{TN}_m  &          &/\; ( \text{TN}_m  &        &+ \text{FP}_m  &          &)           
    \end{alignat}
    \vspace{-4mm}\begin{equation}\vspace{-1mm}
    \,\,\,\,\,\,\,\text{F$_1^m$-score} =\, \text{F1S}_m \,= \frac{2\text{TP}_m}{2\text{TP}_m+\text{FP}_m+\text{FN}_m}    
    \end{equation}
}
for per-class evaluation whereas the unweighted accuracy $\alpha$ and Cohen's $\kappa$
{
    \small
    \vspace{-1.75mm}\begin{align}
    \alpha &= \frac{\textbf{tr}(\bm{\Lambda})}{\sum_{i,j=1}^{M} \Lambda_{i,j}} = \frac{\textbf{tr}(\bm{\Lambda})}{\bm{J}_{1,M}\bm{\Lambda}\bm{J}_{M,1}},  \\
    \kappa &= \frac{ \textbf{tr}(\bm{\Lambda}) \times \bm{J}_{1,M}\bm{\Lambda}\bm{J}_{M,1}  -  \textbf{tr}\big(\bm{\Lambda} \bm{J}_{M} \bm{\Lambda}\big)  }{ \big( \bm{J}_{1,M}\bm{\Lambda}\bm{J}_{M,1} \big)^2 - \textbf{tr}\big(\bm{\Lambda}\bm{J}_{M}\bm{\Lambda}\big)  }
    \end{align}
}
are utilized for the aggregate, multi-class evaluation.

\autoref{fig:vitrajectories} shows the per-class, and per-household performance results. The best result obtained for the voting in \autoref{eq:votinga} is $\alpha = 89.7\%$ while the unweighted majority vote \autoref{eq:votingb} achieved $88.2\%$ of accuracy. It is observed from the figures that microwaves, compact fluorescent lamps, vacuum cleaners, and laptops are the most identifiable loads whereas temperature control devices (such as air conditioners, heaters, fridges, and even fans) are the least.

\autoref{fig:stability} shows further tests of stability of the adopted model. In the first study, we reduce the size of the training dataset and estimate the total performance using the same cross validation technique. In each test case, one building is saved for evaluation, $54r$ buildings for training where $0<r\leq1$ (which is further partitioned into training and validation households), and the remaining $54(1-r)$ are untouched. As expected, the performance of the model degrades notably as the size of the training data decreases.

In the second study, we use the complete training dataset (i.e. $r=1$) but reduce the sampling frequency of the raw measurements. A suitable FIR filter is utilized for each fractional decimation. This study reveals the robustness of our method with respect to the sampling frequency. As observed in the figure, the accuracy slightly drops but is always above 80\% even at 2.5kHz.

In the third study, we test the sensitivity of the model to the phase shift of the extracted segments. In other words, a single period from each measurement with phase $\tau$ is extracted for testing. The common shortest period of all PLAID measurements is one second. Similarly, the figure reveals the robustness of the proposed model to signal variations in steady-state operation, at least for the short common period available in PLAID.

Finally, we repeat all previous experiments while exploiting external knowledge about the list of appliances in each house in reducing the label space for each test case. In other words, if house $h=1$ does not have an air conditioner, the label space for this building becomes $\tilde{\Omega}_1 = \Omega - {\{\text{\textquotesingle AC\textquotesingle}\}}$ and we rather train 45 binary networks. This of course improves the performance where the best-case accuracy reaches $\alpha = 94\%$, but the question is of course about the availability of this knowledge in practice.

\section{Conclusion and future work}
\label{sec:conclusion}

In this work, we introduced an approach of ensemble of neural networks in addressing the appliance identification problem using raw, high-resolution current and voltage measurements. We evaluated our model on the PLAID dataset with the best performance test achieving 89\% of accuracy.

It was observed that the performance notably degrades as the size of the training data is reduced. Therefore, one of the main future work packages is semi-supervised learning methods.

%\input{./chapters/test.tex}

%===============================================================================================================
%=========================================  References/Bibliography  ===========================================
{
\bibliographystyle{./bib/IEEEtranKarim}
\bibliography{./bib/IEEEabrv,./bib/References}
}

\end{document}